\PassOptionsToPackage{table}{xcolor}
\documentclass[11pt]{article}
\usepackage{prove_arxiv}

\title{\texorpdfstring{\textcolor{proposerblue}{Pro}\textcolor{verifierorange}{Ve}: \textcolor{proposerblue}{Proposer}-\textcolor{verifierorange}{Verifier} Decoding}{ProVe: Proposer-Verifier Decoding}}
\author{Ziqi Jin\authormark{\star *}, Weiwen Xu\authormark{*\ddagger}, Jiaxi Li\authormark{*}, Yujie Zhong\authormark{*},
        Qi She\authormark{*}, Aixin Sun\authormark{\star\ddagger}}
\correspondence{\textbf{Corresponding authors:} weiwen.xuu@gmail.com and axsun@ntu.edu.sg}
\authornote{\star}{Nanyang Technological University.}
\authornote{*}{ByteDance.}
\titlenote{Project Page: \url{https://xalp.net/prove}.}
\runningtitle{ProVe: Proposer-Verifier Decoding}
\runninghead{Preprint}

\begin{document}
\maketitle
\begin{abstract}
Speculative decoding accelerates autoregressive (AR) language models by letting a small model draft tokens serially and a large model verify them in parallel.
We reverse these roles.
Starting from a standard AR checkpoint, we fine-tune the large model into a proposer that predicts a block of future tokens in a single forward pass, then use one additional forward pass to verify the proposed block.
This gives \ours{}, proposer-verifier decoding.
The verifier can be a smaller AR model or the proposer itself; the latter gives \lossless{}, which reproduces the proposer's greedy AR decoding.
Block proposals already provide substantial room for acceleration: with a block size of 16, the proposed block has an average AR-matching prefix of 6.6 tokens on GSM8K.
Across six benchmarks with a 32B proposer, \lossless{} reaches $2.5\times$ average AR throughput without training a separate draft head.
With a small 3B verifier, \ours{} reaches $2.7\times$ average throughput at 2.6 accuracy points below AR on the six-task average, providing a controllable tradeoff between speed and accuracy.
\end{abstract}

\noindent\begin{minipage}{\textwidth}
  \centering\captionsetup{hypcap=false}
  \includegraphics[width=\textwidth]{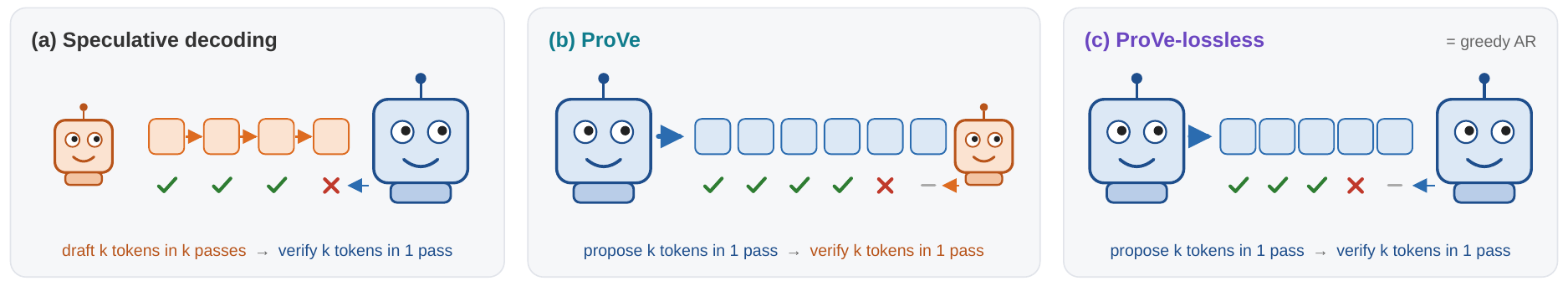}
  \vspace{-10pt}
  \captionof{figure}{Three ways to spend two model calls per decoding cycle. (a)~Speculative decoding: a small model drafts one token per pass and the large model verifies the chain in one pass. (b)~\oursc{}: the large \textcolor{proposerblue}{proposer} writes a proposed block of $B$ tokens in one pass, and a small \textcolor{verifierorange}{verifier} reads it in one more and commits the accepted prefix. (c)~\losslessc{}: the proposer verifies its own proposed block; the committed prefix plus its correction token equals greedy AR decoding, token for token.}
  \label{fig:teaser}
\end{minipage}

\section{Introduction}
\label{sec:intro}
Autoregressive (AR) language models produce one token per forward pass: a response takes as many serial forward passes as it has tokens.
Speculative decoding~\citep{leviathan2023fast, chen2023accelerating} cuts the number of passes of the large model by adding a small model.
A small draft model proposes a sequence of tokens, one at a time. The large target model scores the sequence in one forward pass and keeps the proposed tokens up to the first one it disagrees with.
The gain is set by how well the small model predicts the large one.
Yet the draft model has to be much smaller than the target for its serial passes to be cheap. 
The roles are thus fixed by cost: drafting is serial, so it goes to the cheap model; verifying is one pass, so it goes to the expensive one.

We ask whether the usual assignment of roles is still necessary once the large model can propose multiple tokens in a single forward pass.
\mars{}~\citep{jin2026mars} provides exactly this capability: starting from an ordinary AR checkpoint, it fine-tunes the model into a \emph{block proposer} that predicts the next $B$ tokens in one causal forward pass.
The first proposed token is exactly the model's ordinary next-token prediction, while later positions are predicted without access to the intervening future tokens (Section~\ref{sec:prelim}).
Proposal quality decreases with depth in practice (Figure~\ref{fig:decay}), but the key systems consequence is that proposal is no longer a chain of $B$ serial model calls.

\begin{figure}[t]
\centering
\includegraphics[width=0.76\textwidth]{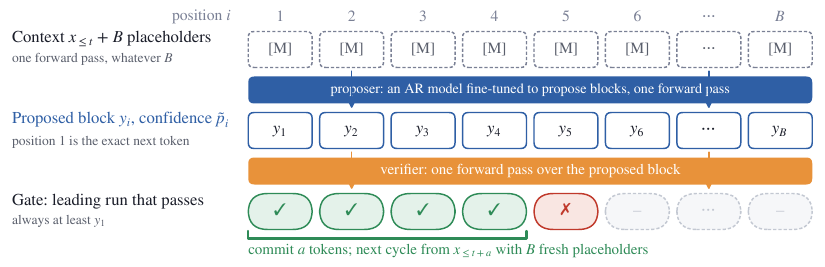}
\caption{One \ours{} cycle. The proposer predicts a block of candidate tokens from placeholders in one forward pass. A second pass verifies the proposed block before the next cycle begins.}
\label{fig:method}
\end{figure}

This changes the role assignment itself.
Once the large model can produce a candidate block in one pass, the small model no longer has to spend serial steps drafting it; instead, it can read the realized block once and judge how much to keep.
This gives \textbf{\ours{}}, \textbf{Pro}poser-\textbf{Ve}rifier decoding (Figure~\ref{fig:method}): the large block-proposing model proposes, and a verifier reads the proposed block in a second pass before deciding how much of it to commit.
In the default \ours{} setting, a smaller AR model serves as the verifier.
In \lossless{}, the proposer verifies its own block, so the resulting sequence reproduces the proposer's greedy AR decoding.
Both variants replace many serial AR steps with exact two forward passes per decoding cycle.

Why pay for a verifier when proposer confidence is already available for free?
\mars{} commits from the same proposed block using the proposer's own confidence, requiring no additional model pass.
This gives a controlled comparison: the proposer, the candidate block, and its confidence are held fixed, and \ours{} changes only what happens after the block has been proposed.
We find that the free confidence rule leaves much of the usable prefix uncommitted: along greedy continuations of GSM8K problems, the proposed block has an average AR-matching prefix of 6.6 tokens, while the default confidence rule commits only 3.0.
Lowering the confidence threshold commits more tokens, but also increases the rate of committing tokens that diverge from greedy AR decoding (Section~\ref{sec:method_gap}).
The central question is therefore whether one additional verification pass can recover enough of this unused prefix to repay its cost.

We evaluate \ours{} and \lossless{} with a 32B instruction proposer over six benchmarks and with a 32B reasoning proposer (Section~\ref{sec:exp}).
\lossless{} reproduces the proposer's greedy AR decoding and reaches $2.5\times$ AR throughput on average, ranging from $1.4\times$ on short constrained text to $4.0\times$ on templated chains of thought.
It outperforms a public EAGLE-3 head and nearly matches a head fine-tuned specifically for the proposer, without training a separate draft head.
With a smaller verifier, \ours{} provides a controllable speed--accuracy tradeoff: a 7B verifier reaches $2.3\times$ average throughput with mean accuracy within 0.3 points of AR, while a 3B verifier reaches $2.7\times$ at 2.6 points below AR.

\begin{contributions}
\begin{itemize}[leftmargin=*,itemsep=2pt,topsep=2pt]
\item Building on the block-proposing capability of \mars{}, we reverse the usual speculative-decoding roles: the large model proposes $B$ tokens in one pass, and a verifier reads the realized block in one more. The verifier can be a smaller AR model or, in \losslessc{}, the proposer itself (Section~\ref{sec:method}).

\item We isolate the value of verification with a controlled comparison to the original \mars{} decoding rule: the proposer and proposed block are identical, while the confidence-only rule is free and \oursc{} deliberately pays one additional pass to read the realized trajectory. On GSM8K, the block has an average AR-matching prefix of 6.6 tokens, while the default confidence rule commits only 3.0 (Section~\ref{sec:method_gap}).

\item At 32B scale, \losslessc{} reproduces the proposer's greedy AR decoding at $2.5\times$ average throughput without a separate draft head, while smaller verifiers provide a tunable speed--accuracy tradeoff reaching $2.7\times$ average throughput with a 3B verifier (Section~\ref{sec:exp_main}).
\end{itemize}
\end{contributions}

\section{Preliminaries: Block-Proposing AR Models}
\label{sec:prelim}

\ours{} builds on a block-proposing AR model: an ordinary decoder-only AR model that is additionally trained to predict the next $B$ tokens in one forward pass.
We obtain this proposer following \mars{}~\citep{jin2026mars}.
Appending $B$ placeholders to the context allows one causal forward pass to produce $B$ predictions for the next $B$ tokens.
When no placeholders are present, the model is decoded as an ordinary AR model.

\begin{figure}[t]
\centering
\includegraphics[width=0.80\textwidth]{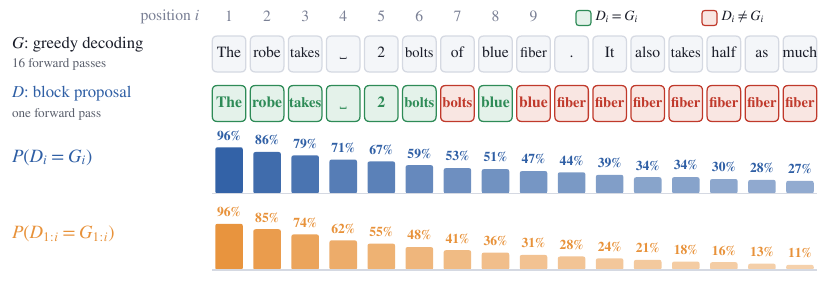}
\caption{
Block proposal quality decays with depth.
Top: an example with $B=16$, comparing the proposer's one-pass block proposal $D$ with its token-by-token greedy continuation $G$.
Bottom: over 437 proposed blocks from 64 GSM8K continuations, both the probability that an individual position matches, $P(D_i=G_i)$, and the probability that the entire prefix through that position matches, $P(D_{1:i}=G_{1:i})$, decrease with position.
}
\label{fig:decay}
\end{figure}

\paragraph{Block proposal.}
Let $P_{\proposer}$ denote the proposer and let \mask{} denote its placeholder token.
Given a context $x_{\le t}$ followed by $B$ copies of \mask{}, a single causal forward pass produces $B$ distributions over the next $B$ tokens:
\begin{equation}
\tilde{p}_i(v)
=
P_{\proposer}\!\left(
v
\,\middle|\,
x_{\le t},
\underbrace{\mask,\ldots,\mask}_{i-1}
\right),
\qquad i=1,\ldots,B.
\label{eq:draft_dist}
\end{equation}
The first distribution, $\tilde{p}_1$, is exactly the proposer's ordinary AR next-token distribution.
The asymmetry across positions is important.
For $i=1$, no future token is missing from the conditioning context.
For $i=2$, the proposer predicts one step further without seeing the token at position $t+1$; for larger $i$, progressively more intervening future tokens are replaced by placeholders.
Deeper positions therefore solve an increasingly difficult prediction problem within the same forward pass.
The proposal is obtained by greedy readout,
$y_i=\arg\max_v \tilde{p}_i(v)$,
and we use $\tilde{p}_i \equiv \tilde{p}_i(y_i)$ to denote the confidence assigned to the proposed token.
As Figure~\ref{fig:decay} shows, proposals are most reliable near the beginning of the block and become less reliable with depth in practice.

\paragraph{Training.}
Following \mars{}~\citep{jin2026mars}, training combines the ordinary next-token objective with a block-prediction objective on placeholder-corrupted responses.
The former preserves ordinary AR generation, while the latter teaches the model to predict multiple future tokens in one pass.
Importantly, block proposal does not introduce a separate draft network or prediction head: ordinary AR prediction and block prediction use the same transformer and output head.
The placeholders only provide the input pattern on which the model learns to predict farther into the future.
Full training details are given in Appendix~\ref{app:impl}.

\paragraph{Original decoding rule.}
\mars{} uses the proposer's own confidence to decide how much of each block to commit.
It always commits the first proposed token, then continues from left to right while the confidence remains above a threshold $\tau$.
The committed prefix is appended to the context, and the proposer generates a fresh block from the new position.
Thus, each cycle uses one forward pass and commits between one and $B$ tokens.
Section~\ref{sec:method_gap} examines how much of the proposed block this confidence rule is able to use.
\section{Proposer-Verifier Decoding}
\label{sec:method}

We take the \mars{} block proposer as fixed and change what happens after it produces a block.
The original \mars{} rule commits from the proposer's confidence, which is already available from the proposal pass.
\ours{} deliberately spends one additional forward pass so that a verifier can read the realized proposed tokens themselves before deciding how much to commit.
This makes \mars{} a particularly clean control: the proposer and candidate block are unchanged, and the only new computation is verification.
We first ask whether this extra pass provides information worth paying for, then describe two choices of verifier.

\subsection{Why Verify the Proposed Block?}
\label{sec:method_gap}

A block proposal is useful until it first diverges from the proposer's greedy AR continuation.
To measure how much useful content the proposer already produces, we run it along its own greedy continuations on randomly selected 300 GSM8K problems and propose a block of 16 tokens at every position.
We compare each proposed block with the continuation obtained by token-by-token greedy decoding.
The average AR-matching prefix is 6.6 tokens.

The original \mars{} rule is a strong baseline precisely because its confidence signal is free: $\tilde{p}_i$ is produced by the same forward pass that produces the candidate $y_i$.
With its default confidence threshold ($\tau{=}0.95$), however, it commits only 3.0 tokens per block on average.
Lowering the threshold commits more tokens, but also increases the frequency of committing a token that diverges from greedy AR decoding.
The proposed block therefore contains useful tokens that cannot be recovered simply by relaxing the free confidence rule.
For an additional verifier pass to be worthwhile, it must identify enough of this unused matching prefix to compensate for its own latency.

The verifier provides an additional signal.
Given the context and a proposed block $y_1,\ldots,y_B$, an AR verifier $V$ assigns each proposed token the probability
\begin{equation}
q_i
=
P_V\!\left(
y_i
\mid
x_{\le t}, y_1,\ldots,y_{i-1}
\right),
\qquad i=1,\ldots,B.
\label{eq:verifier_score}
\end{equation}
All $B$ scores are obtained in one teacher-forced forward pass.
Unlike the proposer's confidence, which is computed from placeholders, the verifier scores each token after reading the actual prefix.

The two scores, $\tilde{p}_i$ and $q_i$,  answer different questions.
The proposer confidence $\tilde{p}_i$ asks whether $y_i$ is plausible at that future position before the intervening future tokens have been instantiated.
The verifier probability $q_i$ asks whether the same token remains plausible after the concrete proposed prefix $y_{<i}$ has been revealed.
Conceptually, $\tilde{p}_i$ plays a prior-like role by exposing the proposer's own preference, while $q_i$ is trajectory-conditioned on the proposal that actually materialized.
A token can therefore look plausible to the proposer in advance yet become inconsistent with the particular trajectory proposed before it.

This also explains why the verifier can be much smaller than the proposer.
The proposer must be capable of predicting several future positions while the intervening tokens remain unknown, whereas the verifier receives those tokens through teacher forcing.
The verifier is not asked to reproduce the proposer's long-horizon prediction ability; it only judges whether a proposed continuation remains locally plausible after the missing context has been filled in.

\begin{figure}[t]
\centering
\includegraphics[width=0.78\textwidth]{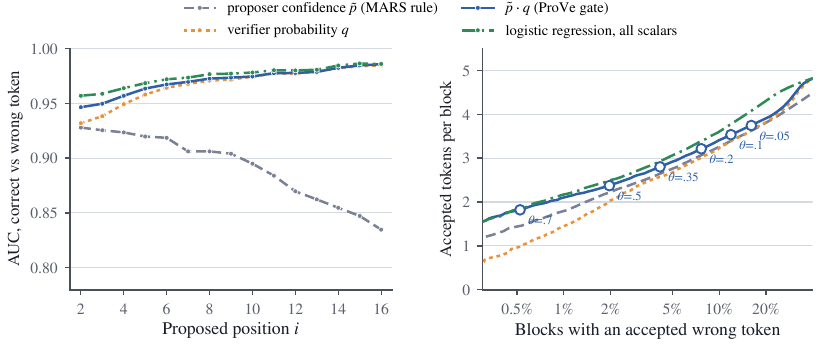}
\caption{Left: AUC for distinguishing proposed tokens that match the greedy AR token, by proposed position with a 3B verifier on GSM8K. Right: committed tokens per block against the fraction of blocks in which the gate commits a non-matching token as each score's threshold is swept.}
\label{fig:gate_depth}
\end{figure}

The proposer confidence is still useful because it records how strongly the proposer itself preferred the candidate.
We therefore treat $\tilde{p}_i\cdot q_i$ as a simple agreement score: it is high only when the candidate is both strongly supported by the proposer and compatible with the realized proposed prefix according to the verifier.
Figure~\ref{fig:gate_depth} compares the two signals and their product on the same proposed blocks.
Verifier probability separates AR-matching from non-matching proposals better than proposer confidence, with AUC $0.960$ versus $0.910$.
Their product reaches an AUC of $0.964$ and performs comparably to the best hand-designed combinations we tested.
Because the product is simple and parameter-free, we use it as the gate score in \ours{}.

\subsection{Verification}
\label{sec:verification}

\ours{} decodes in repeated propose--verify cycles.
Given the current context, the proposer first generates a block of $B$ candidate tokens in one forward pass.
The verifier then reads the context together with the proposed block in one teacher-forced pass.
A prefix of the proposed block is committed, and the next cycle starts from the extended context.
There are two ways to verify the proposed block.

\paragraph{\ours{}: a smaller model verifies.}
Let $\tilde{p}_i$ be the proposer's confidence in its $i$-th proposed token and $q_i$ the verifier probability from Equation~\ref{eq:verifier_score}.
We score each proposed token by their product, $\tilde{p}_i\cdot q_i$.
For a threshold $\theta$, we commit the longest leading prefix whose tokens pass the threshold:
\begin{equation}
a
=
\max
\left\{
t \in \{1,\ldots,B\} :
\tilde{p}_i\cdot q_i \ge \theta
\ \text{for all } 2 \le i \le t
\right\}.
\label{eq:prove_gate}
\end{equation}
The first token is always committed because it is the proposer's ordinary next-token prediction.
A higher threshold is more conservative; lowering it commits longer prefixes but can also commit tokens that diverge from greedy AR decoding.
The verifier never generates tokens itself.
It only decides where to cut the proposed block.

\paragraph{\lossless{}: the proposer verifies itself.}
A special case is to use the proposer as its own verifier.
After proposing a block, we feed the proposed tokens back to the proposer as ordinary tokens in a second forward pass.
This pass gives the proposer's greedy next-token prediction after every proposed prefix.

We compare the proposal with these predictions from left to right.
As long as they agree, the proposed prefix is exactly the prefix that ordinary greedy AR decoding would have produced.
At the first disagreement, we commit the proposer's greedy prediction instead and end the cycle.
If all $B$ proposed tokens agree, we additionally commit the next greedy token produced after the full block.

Every token committed by the cycle therefore lies on the proposer's greedy AR trajectory.
This gives a simple induction argument for exactness: if the context at the start of a cycle is a greedy-AR prefix, matching proposed tokens keep the decoder on that same prefix, and the first disagreement is replaced by exactly the greedy next token.
If the whole block matches, the token predicted after the full block is again the next greedy token.
Repeating the argument across cycles reproduces the proposer's greedy AR decoding token for token.
\lossless{} therefore requires \textit{no threshold and no second model}; each cycle consists of exactly two proposer forward passes.

\subsection{Cost}
\label{sec:cost}

The advantage of reversing the two roles comes from where the serial work is paid.
Let $c_P$ and $c_V$ denote the cost of one proposer and verifier pass, respectively,
and define $\rho=c_V/c_P$.
Normalizing one proposer pass to unit cost, the per-cycle costs $C$ are
\begin{equation}
C_{\text{\ours}} = 1+\rho,
\qquad
C_{\text{\lossless}} = 2,
\qquad
C_{\mathrm{Spec}} = 1+k\rho,
\label{eq:cost}
\end{equation}
where speculative decoding uses the same small model for $k$ serial drafting steps.
If a cycle commits $a$ tokens, its idealized speedup over ordinary AR decoding is $a/C$.

The key difference is the factor multiplying $\rho$.
Speculative decoding invokes the small draft model $k$ times per cycle, whereas
\ours{} invokes the verifier only once.
Increasing the proposal block size can expose more candidate tokens without adding
another serial model call.
A larger verifier can therefore be affordable in regimes where using the same model for $k$ serial draft steps is not.
For example, the 3B model in our implementation has $\rho\approx0.34$.
The idealized per-cycle cost is then about $1.34$ proposer-pass equivalents for \ours{}, but $1+5\rho\approx2.70$ when the same model is used for five serial speculative-drafting steps.
The small model has not become cheaper; its cost is simply paid once per proposed block rather than once per serial draft step.
Actual speedup also depends on how many tokens each method commits per cycle.

Equation~\ref{eq:cost} ignores the small additional cost of processing a block and serving overheads.
Measured pass costs and implementation details are given in Appendix~\ref{app:impl}.

\section{Experiments}
\label{sec:exp}

There are three decoding rules making different use of the same block proposer.
\mars{} uses one proposer pass and decides how far to proceed from the proposer's placeholder-based confidence alone.
\ours{} adds one pass of a smaller model that reads the realized proposed block, yielding a tunable speed--accuracy tradeoff.
\lossless{} instead spends a second proposer pass to recover the exact greedy AR trajectory.
The experiments below evaluate  whether the extra verification pass pays for itself in throughput, and how the choice of verifier changes the resulting operating point.

\paragraph{Setup.}
Our main proposer is Qwen2.5-32B-Instruct~\citep{qwen2024qwen25}, adapted into a block-proposing AR model with $B=16$ following \mars{}~\citep{jin2026mars}.
To isolate the incremental effect of block training, we use a two-stage controlled pipeline: one epoch of ordinary AR SFT on the same instruction data, followed by one epoch of block-proposal training.
The stage-1 AR checkpoint is a control rather than a requirement of \ours{}; in practice, block training can continue from an already post-trained checkpoint using matched in-domain or post-training data.
We use Qwen2.5-3B-Instruct and 7B-Instruct as verifiers, either as stock checkpoints or fine-tuned on generations from the proposer.
Appendix~\ref{app:impl} reports training details and the effect of block training on ordinary AR quality.

\paragraph{Baselines.}
We compare against five decoding baselines: standard autoregressive (AR) decoding, \lossless{}, the original \mars{} confidence rule, EAGLE-3~\citep{li2025eagle3}, and standard speculative decoding.
AR performs ordinary greedy decoding with the same fine-tuned 32B model used as the proposer in our method, without block prediction or verification. We use it as the reference point for both task accuracy and decoding throughput.
For \ours{}, we evaluate confidence thresholds $\theta\in\{0.10,0.05,0.01\}$, where $\theta$ is applied to the combined confidence score $\tilde{p}_i\cdot q_i$.
The original \mars{} confidence rule serves as a no-extra-pass control: it uses the same proposer and produces the same candidate blocks as \ours{}, but directly commits the proposed tokens without an additional verification pass.
For EAGLE-3, we used both its publicly released draft head and a draft head fine-tuned specifically for our 32B proposer.
Finally, we evaluate standard speculative decoding using the same stock 3B or 7B model as a serial draft model, with the 32B model serving as the verifier. This provides a controlled comparison in which the small model is held fixed while its role is reversed relative to our setup.
We use speculative draft lengths $k\in\{3,5\}$ and report sensitivity to $k$ in Appendix~\ref{app:spec-k}.
All methods use greedy decoding.

\paragraph{Evaluation.}
We evaluate on the full test splits of HumanEval, GSM8K, GPQA, BBH, MMLU-Pro, and IFEval using the lm-evaluation-harness~\citep{gao2024lmeval}, with up to 1024 generated tokens.
These benchmarks span short constrained outputs and substantially longer structured continuations, which is useful because block-decoding speed depends on how long a proposed continuation remains predictable.
Accuracy and throughput are measured from the same generations.
All methods run in the same SGLang-based serving stack~\citep{zheng2024sglang}, with a single request in flight and the 32B proposer tensor-parallelized over two GPUs.
Each configuration is compared with its own same-node AR run, and we report throughput relative to that reference.
Additional serving details and evaluation uncertainty are given in Appendices~\ref{app:impl} and~\ref{app:uncertainty}.

\subsection{Main results}
\label{sec:exp_main}

\begin{table}[t]
\centering
\caption{Six general benchmarks, 32B instruction proposer. Exact decoders report relative throughput only, because their algorithmic score change is zero. Lossy cells report score changes relative to greedy AR (percentage points) / throughputs relative to AR. Spec uses the stock 3B or 7B model as its draft ($k=5$); \ours{} uses the same model as its verifier. Absolute task scores are reported in Appendix~\ref{app:absolute-scores}, and evaluation uncertainty in Appendix~\ref{app:uncertainty}. }
\label{tab:main}
\small
\setlength{\tabcolsep}{3pt}
\shrinktofit{%
\begin{tabular}{llrlrlrlrlrlrlrl}
\toprule
Method & Small model & \multicolumn{2}{c}{HumanEval} & \multicolumn{2}{c}{GSM8K} & \multicolumn{2}{c}{GPQA} & \multicolumn{2}{c}{BBH} & \multicolumn{2}{c}{MMLU-Pro} & \multicolumn{2}{c}{IFEval} & \multicolumn{2}{c}{Avg} \\
\midrule
AR (reference) & -- & 82.3 & 68 tok/s & 89.5 & 70 tok/s & 36.2 & 69 tok/s & 84.0 & 69 tok/s & 60.1 & 69 tok/s & 83.2 & 69 tok/s & 72.6 & 1.00$\times$ \\
\midrule
\multicolumn{16}{l}{\cellcolor[HTML]{D6D3D1}\textit{Exact (output equals AR; relative throughput only)}} \\
\lossless{} & -- & \multicolumn{2}{c}{\cellcolor[HTML]{AAC1E4}2.42$\times$} & \multicolumn{2}{c}{\cellcolor[HTML]{8FAEDC}2.86$\times$} & \multicolumn{2}{c}{\cellcolor[HTML]{B9CCE9}2.16$\times$} & \multicolumn{2}{c}{\cellcolor[HTML]{4B7DC6}\textbf{3.98}$\times$} & \multicolumn{2}{c}{\cellcolor[HTML]{ADC3E5}2.37$\times$} & \multicolumn{2}{c}{\cellcolor[HTML]{E5ECF7}1.43$\times$} & \multicolumn{2}{c}{\cellcolor[HTML]{A2BCE2}2.54$\times$} \\
EAGLE-3, chain of 5 & head & \multicolumn{2}{c}{\cellcolor[HTML]{B7CBE8}2.20$\times$} & \multicolumn{2}{c}{\cellcolor[HTML]{CCDAEF}1.84$\times$} & \multicolumn{2}{c}{\cellcolor[HTML]{D1DDF0}1.77$\times$} & \multicolumn{2}{c}{\cellcolor[HTML]{B3C8E7}2.26$\times$} & \multicolumn{2}{c}{\cellcolor[HTML]{C4D4EC}1.98$\times$} & \multicolumn{2}{c}{\cellcolor[HTML]{E9EFF8}1.37$\times$} & \multicolumn{2}{c}{\cellcolor[HTML]{C8D8EE}1.90$\times$} \\
EAGLE-3, tree of 60 & head & \multicolumn{2}{c}{\cellcolor[HTML]{8FAEDC}2.86$\times$} & \multicolumn{2}{c}{\cellcolor[HTML]{AFC5E6}2.33$\times$} & \multicolumn{2}{c}{\cellcolor[HTML]{B8CCE9}2.17$\times$} & \multicolumn{2}{c}{\cellcolor[HTML]{9FBAE1}2.59$\times$} & \multicolumn{2}{c}{\cellcolor[HTML]{ADC4E5}2.36$\times$} & \multicolumn{2}{c}{\cellcolor[HTML]{DBE5F4}1.59$\times$} & \multicolumn{2}{c}{\cellcolor[HTML]{B0C5E6}2.32$\times$} \\
EAGLE-3 fine-tuned, chain of 5 & head & \multicolumn{2}{c}{\cellcolor[HTML]{A9C1E4}2.43$\times$} & \multicolumn{2}{c}{\cellcolor[HTML]{B1C7E6}2.29$\times$} & \multicolumn{2}{c}{\cellcolor[HTML]{BFD1EB}2.06$\times$} & \multicolumn{2}{c}{\cellcolor[HTML]{A8C0E3}2.45$\times$} & \multicolumn{2}{c}{\cellcolor[HTML]{BED0EA}2.08$\times$} & \multicolumn{2}{c}{\cellcolor[HTML]{E4EBF6}1.45$\times$} & \multicolumn{2}{c}{\cellcolor[HTML]{BBCEEA}2.13$\times$} \\
EAGLE-3 fine-tuned, tree of 60 & head & \multicolumn{2}{c}{\cellcolor[HTML]{81A4D7}\textbf{3.08}$\times$} & \multicolumn{2}{c}{\cellcolor[HTML]{87A8D9}\textbf{3.00}$\times$} & \multicolumn{2}{c}{\cellcolor[HTML]{ABC3E5}\textbf{2.38}$\times$} & \multicolumn{2}{c}{\cellcolor[HTML]{90AFDC}2.83$\times$} & \multicolumn{2}{c}{\cellcolor[HTML]{A3BCE2}\textbf{2.52}$\times$} & \multicolumn{2}{c}{\cellcolor[HTML]{D6E1F2}\textbf{1.68}$\times$} & \multicolumn{2}{c}{\cellcolor[HTML]{9FBAE1}\textbf{2.58}$\times$} \\
Spec $k{=}5$ & 3B stock & \multicolumn{2}{c}{\cellcolor[HTML]{CCDAEF}1.85$\times$} & \multicolumn{2}{c}{\cellcolor[HTML]{E3EAF6}1.47$\times$} & \multicolumn{2}{c}{\cellcolor[HTML]{D8E3F3}1.64$\times$} & \multicolumn{2}{c}{\cellcolor[HTML]{BED0EB}2.07$\times$} & \multicolumn{2}{c}{\cellcolor[HTML]{DBE5F4}1.60$\times$} & \multicolumn{2}{c}{\cellcolor[HTML]{F5F8FC}1.16$\times$} & \multicolumn{2}{c}{\cellcolor[HTML]{D9E3F3}1.63$\times$} \\
Spec $k{=}5$ & 7B stock & \multicolumn{2}{c}{\cellcolor[HTML]{DEE7F5}1.54$\times$} & \multicolumn{2}{c}{\cellcolor[HTML]{F2F5FB}1.22$\times$} & \multicolumn{2}{c}{\cellcolor[HTML]{E6EDF7}1.42$\times$} & \multicolumn{2}{c}{\cellcolor[HTML]{D1DDF0}1.77$\times$} & \multicolumn{2}{c}{\cellcolor[HTML]{EBF1F9}1.33$\times$} & \multicolumn{2}{c}{\cellcolor[HTML]{FAFCFD}1.08$\times$} & \multicolumn{2}{c}{\cellcolor[HTML]{E7EEF8}1.39$\times$} \\
\midrule
\multicolumn{16}{l}{\cellcolor[HTML]{D6D3D1}\textit{Lossy ($\Delta$ score $|$ relative throughput)}} \\
\mars{} $\tau{=}0.95$ & -- & \cellcolor[HTML]{FFFEFE}0.0 & \cellcolor[HTML]{EFF3FA}1.27$\times$ & \cellcolor[HTML]{FDFEFE}0.0 & \cellcolor[HTML]{E1EAF6}1.49$\times$ & \cellcolor[HTML]{F3BFBA}$-2.0$ & \cellcolor[HTML]{EEF3FA}1.28$\times$ & \cellcolor[HTML]{F4C6C2}$-1.7$ & \cellcolor[HTML]{A9C1E4}2.43$\times$ & \cellcolor[HTML]{FBECEB}$-0.6$ & \cellcolor[HTML]{E8EEF8}1.38$\times$ & \cellcolor[HTML]{F8DCD9}$-1.1$ & \cellcolor[HTML]{FDFEFE}1.03$\times$ & \cellcolor[HTML]{F9E2E0}$-0.9$ & \cellcolor[HTML]{E2EAF6}1.48$\times$ \\
\mars{} $\tau{=}0.90$ & -- & \cellcolor[HTML]{F7D7D4}$-1.2$ & \cellcolor[HTML]{E8EEF8}1.38$\times$ & \cellcolor[HTML]{FBEDEB}$-0.6$ & \cellcolor[HTML]{D7E2F2}1.66$\times$ & \cellcolor[HTML]{F1B5B0}$-2.3$ & \cellcolor[HTML]{E9EFF8}1.36$\times$ & \cellcolor[HTML]{EEA8A3}$-2.6$ & \cellcolor[HTML]{98B5DF}2.70$\times$ & \cellcolor[HTML]{FBE9E7}$-0.7$ & \cellcolor[HTML]{E1E9F6}1.50$\times$ & \cellcolor[HTML]{F5C8C4}$-1.7$ & \cellcolor[HTML]{FBFCFE}1.07$\times$ & \cellcolor[HTML]{F6CECA}$-1.5$ & \cellcolor[HTML]{DAE4F3}1.61$\times$ \\
\ours{} $\theta{=}0.10$ & 3B stock & \cellcolor[HTML]{FBEBEA}$-0.6$ & \cellcolor[HTML]{A6BFE3}2.47$\times$ & \cellcolor[HTML]{F9FCFB}$+0.2$ & \cellcolor[HTML]{AFC5E6}2.33$\times$ & \cellcolor[HTML]{FCF0EF}$-0.4$ & \cellcolor[HTML]{C4D4EC}1.98$\times$ & \cellcolor[HTML]{F2B9B4}$-2.2$ & \cellcolor[HTML]{618DCD}3.62$\times$ & \cellcolor[HTML]{F7D3D0}$-1.3$ & \cellcolor[HTML]{BCCFEA}2.11$\times$ & \cellcolor[HTML]{F1B6B1}$-2.2$ & \cellcolor[HTML]{F5F8FC}1.17$\times$ & \cellcolor[HTML]{F8DBD8}$-1.1$ & \cellcolor[HTML]{B2C7E7}2.28$\times$ \\
\ours{} $\theta{=}0.05$ & 3B stock & \cellcolor[HTML]{F4C3BF}$-1.8$ & \cellcolor[HTML]{97B4DE}2.73$\times$ & \cellcolor[HTML]{F7D5D2}$-1.3$ & \cellcolor[HTML]{9BB7E0}2.65$\times$ & \cellcolor[HTML]{FEF8F7}$-0.2$ & \cellcolor[HTML]{B8CCE9}2.17$\times$ & \cellcolor[HTML]{F1B4AE}$-2.3$ & \cellcolor[HTML]{5786CA}3.78$\times$ & \cellcolor[HTML]{F6CFCC}$-1.5$ & \cellcolor[HTML]{B3C8E7}2.25$\times$ & \cellcolor[HTML]{EEA49E}$-2.8$ & \cellcolor[HTML]{EAF0F8}1.35$\times$ & \cellcolor[HTML]{F5C9C5}$-1.6$ & \cellcolor[HTML]{A5BEE3}2.49$\times$ \\
\ours{} $\theta{=}0.01$ & 3B stock & \cellcolor[HTML]{F4C3BF}$-1.8$ & \cellcolor[HTML]{90AEDC}\textbf{2.85}$\times$ & \cellcolor[HTML]{ED9E98}$-3.0$ & \cellcolor[HTML]{8DADDB}2.89$\times$ & \cellcolor[HTML]{F8DAD8}$-1.1$ & \cellcolor[HTML]{A4BDE2}\textbf{2.52}$\times$ & \cellcolor[HTML]{E88981}$-3.6$ & \cellcolor[HTML]{4D7EC7}3.95$\times$ & \cellcolor[HTML]{EA928A}$-3.3$ & \cellcolor[HTML]{A1BBE2}2.55$\times$ & \cellcolor[HTML]{EFAAA4}$-2.6$ & \cellcolor[HTML]{DCE6F4}\textbf{1.58}$\times$ & \cellcolor[HTML]{EFABA5}$-2.6$ & \cellcolor[HTML]{97B4DE}2.72$\times$ \\
\ours{} $\theta{=}0.10$ & 3B trained & \cellcolor[HTML]{E5F5ED}$+0.6$ & \cellcolor[HTML]{A0BAE1}2.57$\times$ & \cellcolor[HTML]{F6D2CF}$-1.4$ & \cellcolor[HTML]{8DADDB}2.88$\times$ & \cellcolor[HTML]{E67C73}$-4.0$ & \cellcolor[HTML]{BACDE9}2.15$\times$ & \cellcolor[HTML]{F3C0BB}$-1.9$ & \cellcolor[HTML]{5785CA}3.79$\times$ & \cellcolor[HTML]{F2BDB8}$-2.0$ & \cellcolor[HTML]{B4C9E7}2.25$\times$ & \cellcolor[HTML]{EFAAA4}$-2.6$ & \cellcolor[HTML]{EEF3FA}1.29$\times$ & \cellcolor[HTML]{F3C1BD}$-1.9$ & \cellcolor[HTML]{A5BEE3}2.49$\times$ \\
\ours{} $\theta{=}0.05$ & 3B trained & \cellcolor[HTML]{FFFFFF}0.0 & \cellcolor[HTML]{99B5DF}2.70$\times$ & \cellcolor[HTML]{F6D2CF}$-1.4$ & \cellcolor[HTML]{86A7D9}3.00$\times$ & \cellcolor[HTML]{F1B6B1}$-2.2$ & \cellcolor[HTML]{B5CAE8}2.22$\times$ & \cellcolor[HTML]{EFACA6}$-2.5$ & \cellcolor[HTML]{5382C9}3.86$\times$ & \cellcolor[HTML]{F0B1AC}$-2.4$ & \cellcolor[HTML]{ABC2E5}2.39$\times$ & \cellcolor[HTML]{F1B6B1}$-2.2$ & \cellcolor[HTML]{E9EFF8}1.37$\times$ & \cellcolor[HTML]{F4C4C0}$-1.8$ & \cellcolor[HTML]{9FBAE1}2.59$\times$ \\
\ours{} $\theta{=}0.01$ & 3B trained & \cellcolor[HTML]{F4C3BF}$-1.8$ & \cellcolor[HTML]{95B2DE}2.76$\times$ & \cellcolor[HTML]{E7837A}$-3.8$ & \cellcolor[HTML]{80A3D7}\textbf{3.11}$\times$ & \cellcolor[HTML]{E98A82}$-3.6$ & \cellcolor[HTML]{A5BEE3}2.49$\times$ & \cellcolor[HTML]{E7837A}$-3.8$ & \cellcolor[HTML]{4B7DC6}\textbf{3.99}$\times$ & \cellcolor[HTML]{E67C73}$-4.1$ & \cellcolor[HTML]{9FBAE1}\textbf{2.59}$\times$ & \cellcolor[HTML]{F0B0AB}$-2.4$ & \cellcolor[HTML]{DEE7F5}1.54$\times$ & \cellcolor[HTML]{EB958D}$-3.2$ & \cellcolor[HTML]{96B3DE}\textbf{2.75}$\times$ \\
\ours{} $\theta{=}0.10$ & 7B stock & \cellcolor[HTML]{99D6B8}$+2.4$ & \cellcolor[HTML]{B4C9E7}2.24$\times$ & \cellcolor[HTML]{D6EEE2}$+1.0$ & \cellcolor[HTML]{AFC5E6}2.33$\times$ & \cellcolor[HTML]{E67C73}$-4.0$ & \cellcolor[HTML]{CAD9EE}1.87$\times$ & \cellcolor[HTML]{F8DAD7}$-1.1$ & \cellcolor[HTML]{6791CF}3.52$\times$ & \cellcolor[HTML]{FCEEED}$-0.5$ & \cellcolor[HTML]{C3D4EC}1.99$\times$ & \cellcolor[HTML]{F8DBD8}$-1.1$ & \cellcolor[HTML]{F4F7FB}1.19$\times$ & \cellcolor[HTML]{FCEDEB}$-0.6$ & \cellcolor[HTML]{B7CBE8}2.19$\times$ \\
\ours{} $\theta{=}0.05$ & 7B stock & \cellcolor[HTML]{99D6B8}$+2.4$ & \cellcolor[HTML]{AEC4E5}2.35$\times$ & \cellcolor[HTML]{CFECDE}$+1.1$ & \cellcolor[HTML]{A9C1E4}2.42$\times$ & \cellcolor[HTML]{F9E2E0}$-0.9$ & \cellcolor[HTML]{C1D2EB}2.03$\times$ & \cellcolor[HTML]{F7D4D1}$-1.3$ & \cellcolor[HTML]{628ECE}3.60$\times$ & \cellcolor[HTML]{FAE4E2}$-0.8$ & \cellcolor[HTML]{BBCEEA}2.12$\times$ & \cellcolor[HTML]{F3C2BE}$-1.8$ & \cellcolor[HTML]{EEF2FA}1.29$\times$ & \cellcolor[HTML]{FEF8F7}$-0.2$ & \cellcolor[HTML]{B0C6E6}2.30$\times$ \\
\ours{} $\theta{=}0.01$ & 7B stock & \cellcolor[HTML]{E5F5ED}$+0.6$ & \cellcolor[HTML]{A7BFE3}2.46$\times$ & \cellcolor[HTML]{F9FCFB}$+0.2$ & \cellcolor[HTML]{9FBAE1}2.59$\times$ & \cellcolor[HTML]{E67C73}$-4.5$ & \cellcolor[HTML]{AFC5E6}2.32$\times$ & \cellcolor[HTML]{F3C1BC}$-1.9$ & \cellcolor[HTML]{5B88CB}3.72$\times$ & \cellcolor[HTML]{F5CBC8}$-1.6$ & \cellcolor[HTML]{AFC5E6}2.33$\times$ & \cellcolor[HTML]{F3C2BE}$-1.8$ & \cellcolor[HTML]{E2EAF6}1.48$\times$ & \cellcolor[HTML]{F6CECA}$-1.5$ & \cellcolor[HTML]{A5BEE3}2.48$\times$ \\
\ours{} $\theta{=}0.10$ & 7B trained & \cellcolor[HTML]{B2E0C9}$+1.8$ & \cellcolor[HTML]{A6BEE3}2.48$\times$ & \cellcolor[HTML]{F7D7D5}$-1.2$ & \cellcolor[HTML]{95B2DE}2.76$\times$ & \cellcolor[HTML]{EEA7A1}$-2.7$ & \cellcolor[HTML]{C0D2EB}2.04$\times$ & \cellcolor[HTML]{F5CBC7}$-1.6$ & \cellcolor[HTML]{5D8ACC}3.68$\times$ & \cellcolor[HTML]{F4C3BF}$-1.8$ & \cellcolor[HTML]{B6CAE8}2.20$\times$ & \cellcolor[HTML]{F9E1DF}$-0.9$ & \cellcolor[HTML]{EFF4FA}1.26$\times$ & \cellcolor[HTML]{F8DCDA}$-1.1$ & \cellcolor[HTML]{AAC2E4}2.40$\times$ \\
\ours{} $\theta{=}0.05$ & 7B trained & \cellcolor[HTML]{CCEADB}$+1.2$ & \cellcolor[HTML]{A3BCE2}2.53$\times$ & \cellcolor[HTML]{F8DCDA}$-1.1$ & \cellcolor[HTML]{90AFDC}2.84$\times$ & \cellcolor[HTML]{F2BDB9}$-2.0$ & \cellcolor[HTML]{BACDE9}2.14$\times$ & \cellcolor[HTML]{F0B2AD}$-2.3$ & \cellcolor[HTML]{5886CB}3.76$\times$ & \cellcolor[HTML]{F0B0AA}$-2.4$ & \cellcolor[HTML]{B1C6E6}2.30$\times$ & \cellcolor[HTML]{F0B0AB}$-2.4$ & \cellcolor[HTML]{EAF0F8}1.35$\times$ & \cellcolor[HTML]{F6CECA}$-1.5$ & \cellcolor[HTML]{A5BEE3}2.49$\times$ \\
\ours{} $\theta{=}0.01$ & 7B trained & \cellcolor[HTML]{E5F5ED}$+0.6$ & \cellcolor[HTML]{9DB8E0}2.62$\times$ & \cellcolor[HTML]{EB948D}$-3.3$ & \cellcolor[HTML]{86A8D9}3.00$\times$ & \cellcolor[HTML]{F1B6B1}$-2.2$ & \cellcolor[HTML]{ABC3E5}2.39$\times$ & \cellcolor[HTML]{E98E86}$-3.5$ & \cellcolor[HTML]{5382C9}3.85$\times$ & \cellcolor[HTML]{E8877F}$-3.7$ & \cellcolor[HTML]{A4BDE2}2.50$\times$ & \cellcolor[HTML]{EFAAA4}$-2.6$ & \cellcolor[HTML]{E2EAF6}1.48$\times$ & \cellcolor[HTML]{F0AFAA}$-2.4$ & \cellcolor[HTML]{9CB7E0}2.64$\times$ \\
\bottomrule
\end{tabular}}
\end{table}

\begin{figure}[t]
\centering
\includegraphics[width=0.90\textwidth]{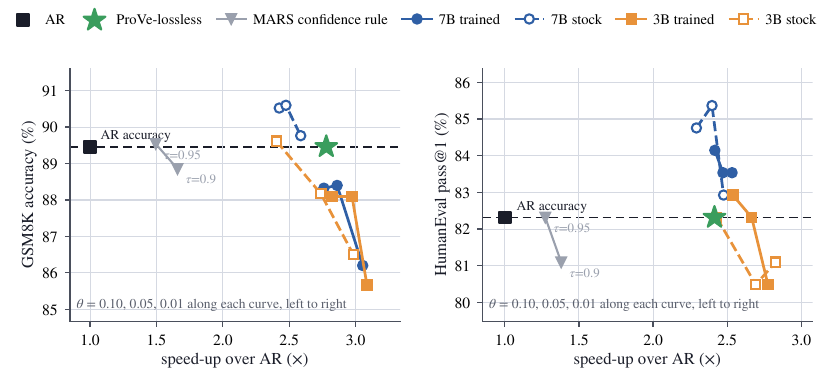}
\caption{Speed against accuracy on the full GSM8K test set (left, $n = 1319$) and HumanEval (right, $n = 164$). The square is AR; the star is \lossless{}, placed on the AR accuracy it reproduces; the grey triangles are the confidence rule of \mars{} at $\tau \in \{0.95, 0.90\}$; each colored curve is \ours{} with one verifier at $\theta \in \{0.10, 0.05, 0.01\}$, dashed for the stock model and solid for the verifier trained on the proposer's outputs. All speedups use same-node AR references; evaluation uncertainty is reported in Appendix~\ref{app:uncertainty}.}
\label{fig:frontier}
\end{figure}
\begin{figure}[t]
\centering
\includegraphics[width=0.80\textwidth]{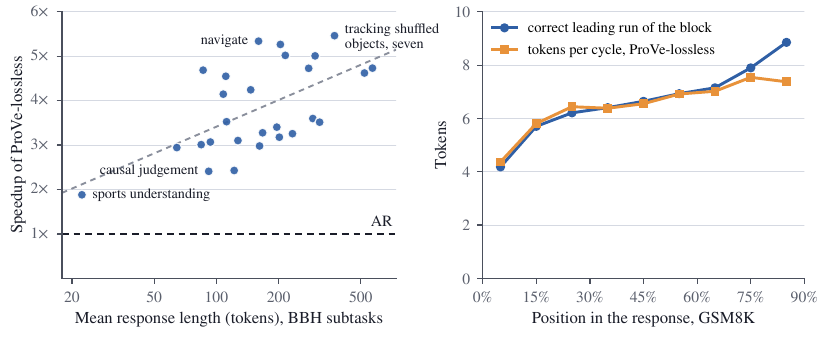}
\caption{Where the gain comes from. Left: \lossless{} speedup on each of the 27 BBH subtasks against the subtask's mean response length (log scale; the same-node runs of Table~\ref{tab:main}), with the fit in log length dashed. Right: a block of 16 proposed at every position of 300 GSM8K greedy continuations, showing the AR-matching prefix length of the proposed block and the tokens \lossless{} commits per cycle when replayed along the trace, by relative position in the response.}
\label{fig:insight}
\end{figure}

\paragraph{\lossless{} provides strong exact acceleration.}
Table~\ref{tab:main} shows that \lossless{} reaches $2.54\times$ AR throughput on average while reproducing the proposer's greedy AR decoding.
The speedup ranges from $1.43\times$ on IFEval to $3.98\times$ on BBH.
The wide range already suggests that acceleration is controlled by the structure of the generated continuation rather than by a fixed model-side speedup; Figure~\ref{fig:insight} returns to this variation below.
\lossless{} requires neither a second model nor a separately trained draft head.
For comparison, the public EAGLE-3 tree reaches $2.32\times$ on average, while an EAGLE-3 head fine-tuned specifically for our proposer reaches $2.58\times$.
Thus, self-verification provides acceleration comparable to an external draft head tahat explicitly trained on proposer's output.

\paragraph{\ours{} trades accuracy for additional speed.}
Using a smaller verifier introduces a controllable speed--accuracy tradeoff through the threshold $\theta$ (Figure~\ref{fig:frontier}).
With the stock 7B verifier at $\theta=0.05$, \ours{} reaches $2.30\times$ average throughput at only $-0.3$ points relative to AR.
Lowering the threshold commits longer prefixes and increases speed: with a stock 3B verifier at $\theta=0.01$, \ours{} reaches $2.72\times$ average throughput at $-2.6$ points.
On GSM8K, the same configuration reaches $2.89\times$ throughput at $-3.0$ points relative to AR.
\lossless{} provides an exact reference point at $2.54\times$ average throughput.

\paragraph{Fine-tuning the verifier improves acceptance efficiency.}
Fine-tuning the verifier on generations from the proposer changes not only its ranking of proposed tokens but also the scale of its probabilities.
A fixed value of $\theta$ can therefore correspond to different levels of conservativeness before and after fine-tuning.
To separate ranking quality from this calibration shift, we compare stock and trained verifiers at matched rates of committing a non-matching token.
At matched error rates, the fine-tuned verifiers commit $6$--$10\%$ more tokens per cycle than their stock counterparts.
We provide the full comparison in Appendix~\ref{app:gate}.

\paragraph{\ours{} avoids serial drafting cost.}
At the conservative $\theta=0.10$ operating point, \ours{} is faster than speculative decoding with the same small model.
With the stock 3B model, speculative decoding reaches $1.63\times$ average throughput, while \ours{} reaches $2.28\times$ at $-1.2$ points relative to AR.
With the stock 7B model, the corresponding throughputs are $1.39\times$ and $2.19\times$, with \ours{} at $-0.6$ points.
Sweeping the speculative draft length over $k\in\{3,5\}$ does not change this comparison; Appendix~\ref{app:spec-k} reports the full sensitivity.
The difference is largest on longer structured generations such as BBH and GSM8K.
This matches the cost analysis in Section~\ref{sec:cost}: speculative decoding calls the small model repeatedly to draft tokens, whereas \ours{} calls it once to verify one proposed block.

\paragraph{Longer responses are usually faster.}
Figure~\ref{fig:insight} shows that the speedup tracks how much of a proposed block can be committed.
Across the 27 BBH subtasks, \lossless{} speedup increases from about $1.9\times$ on the shortest responses to $5.5\times$ on the longest, with a correlation of $r=0.59$ with log response length.
The same mechanism appears within individual GSM8K responses: the average AR-matching prefix grows from 4.2 tokens near the beginning of a response to 8.9 tokens near the end, while \lossless{} commits 4.4 to 7.5 tokens per cycle.
The method therefore benefits most when the continuation becomes predictable enough for a single proposal to match greedy AR for several positions.

\subsection{A reasoning proposer}
\label{sec:exp_reasoning}

We next test whether the same behavior extends to long chain-of-thought generation.
This setting is particularly informative because long reasoning traces provide more opportunities for multi-token acceleration, while decision points can also make future tokens less predictable.
We use Qwen3-32B in thinking mode as the proposer and the stock Qwen3-0.6B as the verifier; the same 0.6B model serves as the draft model for speculative decoding. See Appendix~\ref{app:impl} for proposer training details.

On GSM8K, \lossless{} reaches $2.47\times$ AR throughput, while \ours{} at $\theta=0.10$ reaches $2.34\times$ at $-0.4$ points relative to AR.
On MATH-500, \lossless{} reaches $2.79\times$, and \ours{} at $\theta=0.10$ reaches $3.16\times$ at $-0.2$ points.
Speculative decoding with the same 0.6B model reaches $2.06\times$ and $2.28\times$ on the two benchmarks, respectively.
At $\theta=0.05$, \ours{} roughly matches \lossless{} in throughput on GSM8K at $-1.2$ points and reaches $3.39\times$ on MATH-500 at $-3.8$ points.

\begin{table}[t]
\centering
\caption{Reasoning proposer on GSM8K and MATH-500. Lossy cells report score change relative to greedy AR (percentage points) / throughput relative to AR; exact decoders show relative throughput only. 
Absolute scores are reported in Appendix~\ref{app:absolute-scores}, and evaluation uncertainty in Appendix~\ref{app:uncertainty}.}
\label{tab:gsm8k}
\small
\setlength{\tabcolsep}{4pt}
\shrinktofit[0.55\linewidth]{%
\begin{tabular}{lrlrl}
\toprule
Configuration & \multicolumn{2}{c}{GSM8K (1319)} & \multicolumn{2}{c}{MATH-500 (500)} \\
\midrule
AR (reference) & 94.9 & 73.2 tok/s & 91.6 & 70.4 tok/s \\
\midrule
\multicolumn{5}{l}{\textit{Exact (relative throughput only)}} \\
Spec, $k=5$ & \multicolumn{2}{c}{\cellcolor[HTML]{BED0EA}2.06$\times$} & \multicolumn{2}{c}{\cellcolor[HTML]{B1C6E6}2.28$\times$} \\
\lossless{} & \multicolumn{2}{c}{\cellcolor[HTML]{A6BEE3}\textbf{2.47}$\times$} & \multicolumn{2}{c}{\cellcolor[HTML]{93B0DD}\textbf{2.79}$\times$} \\
\midrule
\multicolumn{5}{l}{\textit{Lossy ($\Delta$ score $|$ relative throughput)}} \\
\ours{}, $\theta=0.10$ & \cellcolor[HTML]{FCF2F1}$-0.4$ & \cellcolor[HTML]{AEC4E5}2.34$\times$ & \cellcolor[HTML]{FEF8F8}$-0.2$ & \cellcolor[HTML]{7CA0D5}3.16$\times$ \\
\ours{}, $\theta=0.05$ & \cellcolor[HTML]{F8D8D5}$-1.2$ & \cellcolor[HTML]{A7BFE3}\textbf{2.45}$\times$ & \cellcolor[HTML]{E7837A}$-3.8$ & \cellcolor[HTML]{6E96D1}\textbf{3.39}$\times$ \\
\bottomrule
\end{tabular}}
\end{table}

\section{Related Work}
\label{sec:relatedwork}

\paragraph{Speculative decoding and draft heads.}
Speculative decoding drafts $k$ tokens with a cheap model and verifies them with the target in one pass, preserving the target distribution~\citep{leviathan2023fast, chen2023accelerating}; later work cuts the draft cost with heads trained on the target's hidden states and tree-structured candidates~\citep{li2024eagle, li2024eagle2, li2025eagle3}, the target's own early layers~\citep{zhang2024draft, elhoushi2024layerskip}, Jacobi-style $n$-gram candidates~\citep{fu2024lookahead} or of a suffix automaton~\citep{hu2025sam}.
Nearly all of this work is exact, the draft only changes speed, and the roles are fixed: drafter are designed to be small and cheap, while the verifier are large and expensive.
Judge Decoding~\citep{bachmann2025judge} and Big Little Decoder~\citep{kim2023bild} relax the verifier's rule, as \ours{} does with its threshold. Neither lets the large model propose.

\paragraph{Block proposals.}
Several methods let a model propose several positions in one of its own passes: PaSS~\citep{monea2023pass} appends learned look-ahead embeddings, Parallel Prompt Decoding~\citep{chen2025ppd} trained prompt tokens, Medusa~\citep{cai2024medusa} extra heads, multi-token prediction~\citep{gloeckle2024better, deepseek2024deepseekv3} extra output modules, and Set Block Decoding~\citep{gat2025set} and \mars{}~\citep{jin2026mars} fine-tune the model to predict masked future tokens, the proposer of this paper.
ReDrafter~\citep{cheng2024recurrent}, Amphista~\citep{li2025amphista}, FastEagle~\citep{huang2025fasteagle} and DFlash~\citep{chen2026dflash} draft a block in one pass of a trained \emph{small} draft network under exact verification by the target, the standard direction with a block draft; IDLM~\citep{yu2026introspective} fuses the verification of one proposed block and the proposal of the next into one pass of the same model.
Block diffusion models~\citep{arriola2025block}, DSpark~\citep{cheng2026dspark} and \mars{} commit on the model's own confidence; what \ours{} adds is the verifying pass that reads the proposed block. And through experiments, we show that the verification step is essential to commit more tokens.

\section{Conclusion}
\label{sec:conclusion}

In this work, we show that block proposals contain substantially more usable content than the proposer's own confidence suggests: on GSM8K, the proposed block has an average AR-matching prefix of 6.6 tokens, while the original confidence rule commits only 3.0. The gap arises because proposer confidence is computed from placeholders and therefore does not see the realized proposed prefix.
A verifier instead reads the proposed tokens themselves, providing a second signal for deciding how much of the block to commit.
This simple change gives \ours{} two useful operating alternatives.
With a smaller verifier, \ours{} provides a controllable speed--accuracy tradeoff; when the proposer verifies itself, \lossless{} reproduces greedy AR decoding with no second model and reaches $2.5\times$ average AR throughput.
Our results suggest that block proposal and verification need not follow the traditional speculative-decoding division of labor.
A large model can propose several tokens in parallel, while verification can be performed in one additional pass.

\bibliographystyle{plainnat}
\bibliography{references}

\appendix
\section{Implementation Details}
\label{app:impl}

\paragraph{Serving.}
All configurations run in one SGLang-based stack~\citep{zheng2024sglang} extended with the \ours{} algorithms, on two data-center GPUs with tensor parallelism of two for the 32B model, bf16 weights, the FlashInfer attention backend and a single request in flight; the AR controls use the unmodified server.
\ours{} is a decoding algorithm with a small configuration: block size 16, the product rule of Equation~\ref{eq:prove_gate} with its threshold, the verifier checkpoint and the fraction of device memory reserved for the in-process verifier (its weights, KV cache and CUDA graphs): 0.18 for the 3B and 0.32 for the 7B, with the proposer's static pool at 0.70 and 0.55 so that the two never compete (an undersized verifier pool does not fail but silently degrades toward one accepted token per cycle; Appendix~\ref{app:validity}).
Each cycle is one scheduler step per pass, and proposed tokens that are not committed are dropped from the proposer's KV cache before the next cycle, so the cache after a cycle equals what AR decoding of the committed context would have produced.
Spec uses the stack's standalone-draft implementation with the same 3B or 7B checkpoint; we evaluate serial draft lengths $k\in\{3,5\}$ and report the sweep below.
EAGLE-3 uses the stack's implementation with the head as a chain of five draft steps and with its default tree of 60 draft tokens (Table~\ref{tab:main}).

\paragraph{Measured pass costs.}
On the same two GPUs (60 GSM8K problems): one AR decode step 14.8 ms; one pass of the 32B over a 16-token window 15.6 ms, both as the proposing and as the verifying pass of \lossless{}; one in-process pass of the 3B verifier 5.0 ms ($\rho = 0.34$) and of the 7B 5.8 ms ($\rho = 0.40$).
The verifier's pass is dominated by a fixed per-call cost of about 2.7 ms that does not depend on its size (the 0.5B costs 3.0 ms, halving the 7B's weight bytes with fp8 saves 0.2 ms, and running it without tensor parallelism is slower, 6.9 ms), which is why a stronger verifier is nearly free in Equation~\ref{eq:cost}.
Per-cycle scheduling and cache-trimming overhead is 8.7 ms in \lossless{} and 4 ms in \ours{}, the part the equation omits; with it, $7.27 / (2 \times 1.05 + 0.59) = 2.70\times$ against the measured $2.70\times$ for \lossless{}, and $4.13 / (1.05 + 0.34 + 0.27) = 2.49\times$ against $2.49\times$ for \ours{} with the 3B verifier at $\theta = 0.10$.

\paragraph{Block size.}
The proposer was trained with 16 placeholders but proposes a longer block when asked.
On the full GSM8K test set \lossless{} commits 5.9 / 7.5 / 8.1 / 8.6 tokens per cycle at $B = 8 / 16 / 24 / 32$ and decodes at $2.44 / 2.79 / 2.84 / 2.84\times$: the speedup saturates from $B = 24$ on because both passes pay for every row of the window, and every other experiment uses the $B = 16$ the proposer was trained with.

\paragraph{The block objective and training.}
With $t_b$ the position before block $b$ and $\tilde{p}^{(b)}_i$ Equation~\ref{eq:draft_dist} evaluated with the context $x_{\le t_b}$, the loss is
\begin{equation}
\mathcal{L} \;=\; -\sum_{j} \log P_{\proposer}\!\left(x_{j+1} \mid x_{\le j}\right) \;-\; \sum_{b} \sum_{i=1}^{B} \log \tilde{p}^{(b)}_i\!\left(x_{t_b+i}\right),
\label{eq:train}
\end{equation}
both streams in one forward pass over the concatenation of the placeholder stream and the original, with an attention pattern in which a placeholder sees the original tokens before its block and the placeholders before it; attention is causal everywhere.
Both proposers use the reference implementation of~\citet{jin2026mars} on 32 GPUs: the instruction proposer 150 GPU hours for the AR stage and 384 for the block stage at 512 tokens per stream; the reasoning proposer 1,194 GPU hours at 16k tokens, a single-benchmark check because reasoning traces are long and block training doubles every sequence.
The trained verifiers are the stock 3B and 7B fine-tuned for two epochs on the proposer's generations (about 4 and 10 GPU hours).

\paragraph{Effect of block training on ordinary AR quality.}
\label{app:adaptation}
The instruction proposer is trained in two stages so that the incremental effect of the block objective can be measured cleanly.
Stage 1 performs one epoch of ordinary AR SFT; stage 2 applies the \mars{} block-proposal objective on the same instruction data.
Table~\ref{tab:ar-retention} evaluates all three checkpoints with ordinary greedy AR decoding.
The controlled comparison is therefore the stage-1 AR checkpoint against the final proposer, not the original Qwen2.5 checkpoint against the final proposer.
Across the six benchmarks, block training changes the unweighted average from 73.1 to 72.5, a difference of $-0.6$ points; every per-task change is within 1.8 points.

\begin{table}[ht]
\centering
\caption{Ordinary greedy-AR accuracy before and after block-proposal training. The stage-1 AR-SFT checkpoint is the controlled reference for isolating the effect of the subsequent \mars{} block-training stage; the original Qwen2.5-32B-Instruct checkpoint is shown for context. Avg is the unweighted mean of the six tasks.}
\label{tab:ar-retention}
\small
\setlength{\tabcolsep}{4pt}
\shrinktofit{%
\begin{tabular}{lrrrrrrr}
\toprule
Checkpoint & HumanEval & GSM8K & GPQA & BBH & MMLU-Pro & IFEval & Avg \\
\midrule
Original Qwen2.5-32B-Instruct & 89.6 & 83.4 & 39.5 & 40.8 & 69.3 & 79.1 & 67.0 \\
AR SFT (stage 1)              & 84.1 & 90.2 & 36.4 & 84.3 & 61.2 & 82.3 & 73.1 \\
Final block proposer          & 82.3 & 89.5 & 36.2 & 84.0 & 60.1 & 83.2& 72.6 \\
\bottomrule
\end{tabular}}
\end{table}

The larger differences between the original checkpoint and stage 1 primarily reflect the ordinary SFT stage and its change in generation behavior, rather than the block objective.
For BBH, the original checkpoint frequently emits paragraph breaks while the evaluation configuration treats \texttt{\textbackslash n\textbackslash n} as a stopping string; using EOS alone raises its score from 40.8 to 62.5, which is still below the stage-1 score of 84.3.
On MMLU-Pro, stage-1 SFT reduces the mean generated length from 372 to 164 tokens, showing that the SFT data substantially changes response style before block training begins.
We use stage 1 only to obtain a controlled reference under matched data; it is not an algorithmic requirement of \ours{}.
In practice, block-proposal training can be applied after an existing post-training stage using matched in-domain or post-training data.

\paragraph{Speculative draft length.}
\label{app:spec-k}
We also vary the number of serial draft steps in speculative decoding, using the same stock 3B and 7B checkpoints as in the main comparison.
Table~\ref{tab:spec-k} shows that neither $k=3$ nor $k=5$ dominates across tasks.
With the 3B draft, $k=5$ has the higher six-task average (1.63$\times$ versus 1.58$\times$), whereas the 7B draft is essentially tied (1.38$\times$ versus 1.39$\times$).
Short constrained generations such as IFEval favor the shorter draft, while BBH benefits from the longer draft.
The main comparison reports $k=5$; the qualitative gap to \ours{} is unchanged under $k=3$.

\begin{table}[ht]
\centering
\caption{Sensitivity of speculative decoding to serial draft length $k$. The stock 3B or 7B checkpoint is used as the draft model; entries are throughput relative to the same-node AR reference.}
\label{tab:spec-k}
\small
\setlength{\tabcolsep}{4pt}
\shrinktofit{%
\begin{tabular}{llrrrrrrr}
\toprule
Draft model & $k$ & HumanEval & GSM8K & GPQA & BBH & MMLU-Pro & IFEval & Avg \\
\midrule
3B stock & 3 & 1.69$\times$ & 1.52$\times$ & 1.56$\times$ & 1.85$\times$ & 1.55$\times$ & 1.28$\times$ & 1.58$\times$ \\
3B stock & 5 & 1.78$\times$ & 1.56$\times$ & 1.60$\times$ & 2.07$\times$ & 1.60$\times$ & 1.18$\times$ & 1.63$\times$ \\
\midrule
7B stock & 3 & 1.48$\times$ & 1.30$\times$ & 1.39$\times$ & 1.65$\times$ & 1.37$\times$ & 1.17$\times$ & 1.39$\times$ \\
7B stock & 5 & 1.49$\times$ & 1.25$\times$ & 1.37$\times$ & 1.78$\times$ & 1.34$\times$ & 1.06$\times$ & 1.38$\times$ \\
\bottomrule
\end{tabular}}
\end{table}

\paragraph{Harness.}
Accuracy is scored by the lm-evaluation-harness fork of~\citet{jin2026mars} with the chat template applied, through a backend that sends one request at a time and records generated tokens and wall-clock time per request; throughput is total tokens over total in-flight seconds of a unit.
Every server passes a fixed-prompt speed probe before use, or is restarted; models are staged to node-local disk before the server starts.

\section{Absolute Evaluation Scores}
\label{app:absolute-scores}

The main tables emphasize direct comparison by reporting score changes relative to the corresponding greedy-AR reference.
For completeness, Tables~\ref{tab:absolute-main} and~\ref{tab:absolute-reasoning} report the underlying absolute task scores.
Positive deltas in the main tables indicate direction only; evaluation uncertainty is reported in Appendix~\ref{app:uncertainty}.

\begin{table}[ht]
\centering
\caption{Absolute task scores underlying the lossy block of Table~\ref{tab:main}. Avg is the unweighted mean of the six tasks. Exact decoders are omitted because their algorithmic target is the AR reference.}
\label{tab:absolute-main}
\small
\setlength{\tabcolsep}{4pt}
\shrinktofit{%
\begin{tabular}{llrrrrrrr}
\toprule
Method & Small model & HumanEval & GSM8K & GPQA & BBH & MMLU-Pro & IFEval & Avg \\
\midrule
AR (reference) & -- & 82.3 & 89.5 & 36.2 & 84.0 & 60.1 & 83.2 & 72.6 \\
\midrule
\mars{} $\tau=0.95$ & none & 82.3 & 89.5 & 34.2 & 82.3 & 59.5 & 82.1 & 71.6 \\
\mars{} $\tau=0.90$ & none & 81.1 & 88.9 & 33.9 & 81.4 & 59.4 & 81.5 & 71.0 \\
\ours{} $\theta=0.10$ & 3B stock & 81.7 & 89.6 & 35.7 & 81.9 & 58.7 & 81.0 & 71.4 \\
\ours{} $\theta=0.05$ & 3B stock & 80.5 & 88.2 & 35.9 & 81.7 & 58.6 & 80.4 & 70.9 \\
\ours{} $\theta=0.01$ & 3B stock & 80.5 & 86.5 & 35.0 & 80.4 & 56.7 & 80.6 & 70.0 \\
\ours{} $\theta=0.10$ & 3B trained & 82.9 & 88.1 & 32.1 & 82.1 & 58.1 & 80.6 & 70.7 \\
\ours{} $\theta=0.05$ & 3B trained & 82.3 & 88.1 & 33.9 & 81.5 & 57.7 & 81.0 & 70.8 \\
\ours{} $\theta=0.01$ & 3B trained & 80.5 & 85.7 & 32.6 & 80.2 & 56.0 & 80.8 & 69.3 \\
\ours{} $\theta=0.10$ & 7B stock & 84.8 & 90.4 & 32.1 & 82.9 & 59.6 & 82.1 & 72.0 \\
\ours{} $\theta=0.05$ & 7B stock & 84.8 & 90.6 & 35.3 & 82.7 & 59.2 & 81.3 & 72.3 \\
\ours{} $\theta=0.01$ & 7B stock & 82.9 & 89.6 & 31.7 & 82.1 & 58.5 & 81.3 & 71.0 \\
\ours{} $\theta=0.10$ & 7B trained & 84.1 & 88.2 & 33.5 & 82.4 & 58.3 & 82.3 & 71.5 \\
\ours{} $\theta=0.05$ & 7B trained & 83.5 & 88.4 & 34.2 & 81.7 & 57.7 & 80.8 & 71.0 \\
\ours{} $\theta=0.01$ & 7B trained & 82.9 & 86.2 & 33.9 & 80.6 & 56.4 & 80.6 & 70.1 \\
\bottomrule
\end{tabular}}
\end{table}

\begin{table}[ht]
\centering
\caption{Absolute scores underlying Table~\ref{tab:gsm8k}. The measured \lossless{} score can differ slightly from AR under bf16 execution even though the decoding rule is algorithmically exact; Appendix~\ref{app:validity} analyzes this effect.}
\label{tab:absolute-reasoning}
\small
\setlength{\tabcolsep}{8pt}
\begin{tabular}{lrr}
\toprule
Configuration & GSM8K & MATH-500 \\
\midrule
AR, proposer (reference) & 94.9 & 91.6 \\
Spec, $k=5$ & 94.3 & 92.2 \\
\lossless{} & 94.0 & 91.6 \\
\ours{}, $\theta=0.10$ & 94.5 & 91.4 \\
\ours{}, $\theta=0.05$ & 93.7 & 87.8 \\
\bottomrule
\end{tabular}
\end{table}

\section{Gating Rules Compared Offline}
\label{app:gate}

\begin{figure}[ht]
\centering
\includegraphics[width=0.92\textwidth]{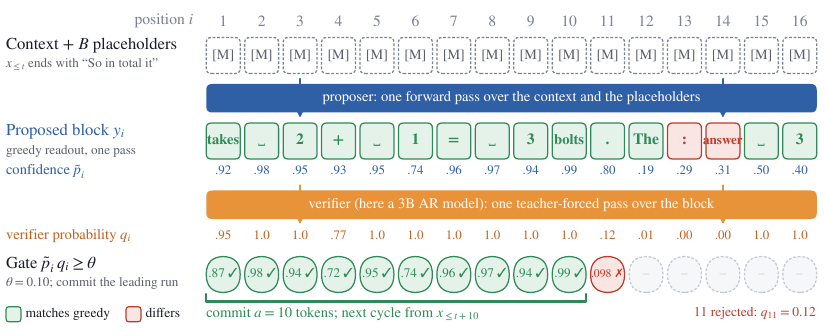}
\caption{One \ours{} cycle on a real block from a GSM8K answer (32B instruction proposer, 3B verifier). Green proposed tokens equal what the proposer generates one token at a time, red ones differ. The gate $\tilde{p}_i \cdot q_i \ge \theta$ with $\theta = 0.10$ commits ten tokens; the eleventh is correct but the verifier gives it 0.12, so the cycle stops there. The confidence rule of \mars{} at $\tau = 0.95$ would commit two, because $\tilde{p}_3 = 0.946$.}
\label{fig:example}
\end{figure}

The study of Section~\ref{sec:method_gap} teacher-forces the proposer along its own greedy continuations of the first 300 GSM8K problems, proposes a block of 16 at every position (26,407 blocks), and records for every proposed token the proposer's top-two probabilities and entropy, the verifier's probability, top-two and entropy, and whether the token equals the greedy token.
A sequential gate only judges tokens whose shallower siblings were correct, so statistics are compared on that population, by the AUC of ranking correct against wrong tokens and by the frontier a gate traces as its threshold is swept: accepted tokens per block against the fraction of blocks in which a wrong token was accepted (Table~\ref{tab:gate}, Figure~\ref{fig:gate_depth}).
The proposer's own confidence loses discrimination as the proposed block deepens, because it is computed under placeholder conditioning, and the product gains it, because the verifier sees the proposed tokens before the one it judges; the deep positions are where the accepted length is decided.
Reweighting the factors, $\tilde{p}_i^{\,\alpha} q_i^{\,\beta}$, is temperature scaling in the log domain and moves the frontier by less than the noise between repeats; a logistic regression on all recorded scalars accepts 2 to 4\% more tokens at matched divergence, the room left for a better scalar gate, which we do not deploy because the product has no parameters.
Figure~\ref{fig:gate_depth} in the main text plots the AUC by position and the swept frontier.

\begin{table}[ht]
\centering
\caption{Gating rules on the population a sequential gate sees (3B verifier, GSM8K, 26,407 blocks). AUC ranks correct against wrong proposed tokens; the remaining columns give accepted tokens per block, beyond the always-committed first token, at a matched fraction of blocks in which a wrong token was accepted.}
\label{tab:gate}
\small
\shrinktofit[0.85\linewidth]{%
\begin{tabular}{lrrrrr}
\toprule
Statistic & AUC & 0.5\% & 1.0\% & 2.0\% & 5.0\% \\
\midrule
proposer confidence $\tilde{p}$ only (\mars{}) & 0.909 & 2.44 & 2.77 & 3.21 & 3.73 \\
verifier probability $q$ only & 0.960 & 1.93 & 2.42 & 3.02 & 3.63 \\
$\tilde{p}\cdot q$ (Equation~\ref{eq:prove_gate}) & 0.964 & 2.78 & 3.10 & 3.37 & 3.92 \\
$\tilde{p}^{2}\cdot q$ & 0.964 & 2.79 & 3.09 & 3.45 & 3.97 \\
$(\tilde{p}_{(1)}-\tilde{p}_{(2)})\cdot q$ & 0.961 & 2.77 & 3.06 & 3.40 & 3.97 \\
$\min(\tilde{p}, q)$ & 0.957 & 2.71 & 2.93 & 3.28 & 3.74 \\
logistic regression on all scalars (held-out prompts) & 0.970 & 2.81 & 3.17 & 3.50 & 4.04 \\
\bottomrule
\end{tabular}}
\end{table}

\section{Measurement Validity}
\label{app:validity}

\paragraph{Exactness of \lossless{}.}
Feeding each generation of the main campaign back to the plain AR server with per-token log probabilities, the tokens of \lossless{} are the argmax at every position except 0.84\%, against 0.85\% for the AR server's own generations scored the same way, with the same distribution of gaps to the argmax (median 0, maximum 0.4 nats): the two decoders differ only in bf16 tie-breaking.
On the reasoning line the generations are long enough for tie-breaking to show: over 300 GSM8K problems, 20 generations are identical token for token, the first differing token sits at a median position of 176, and 266 of the 280 diverged pairs end at the same final answer.

\paragraph{Repeats and pooling.}
Repeats of the same configuration on different nodes agree in accuracy to the item and in throughput within 2\%, except at the loosest thresholds of the 3B and 7B verifiers, which moved by 8 to 10\% in opposite directions; every comparison in the paper is therefore decoded on one node in one job with its own AR run.
Throughput is total generated tokens over total in-flight seconds of a unit; averaging per-request rates would overweight short generations, whose rate is dominated by prefill and the first cycle.

\paragraph{Evaluation uncertainty.}
\label{app:uncertainty}

We evaluate all methods on the full evaluation split of each benchmark.
The test-set sizes and corresponding 95\% confidence half-widths for the main six-benchmark evaluation are:
HumanEval, $n=164$ ($\pm5.9$ points);
GSM8K, $n=1319$ ($\pm1.7$);
GPQA, $n=448$ ($\pm4.4$);
BBH, $n=6511$ ($\pm0.9$);
MMLU-Pro, $n=12032$ ($\pm0.9$);
and IFEval, $n=541$ ($\pm3.2$).
We therefore do not interpret accuracy differences within these ranges as meaningful differences between decoding methods. For the reasoning-model evaluation, MATH-500 contains 500 problems and has a 95\% confidence half-width of approximately $\pm2.4$ points, which covers the accuracy differences among the configurations we report.

\paragraph{A silent failure.}
An undersized in-process verifier pool does not fail but degrades \ours{} toward one accepted token per cycle; with the 7B verifier a fraction of 0.25 starved the pool (probe 119 instead of 172 tok/s) and the campaign uses 0.32 with the proposer's pool at 0.55.

\begin{figure}[ht]
\centering
\includegraphics[width=0.90\textwidth]{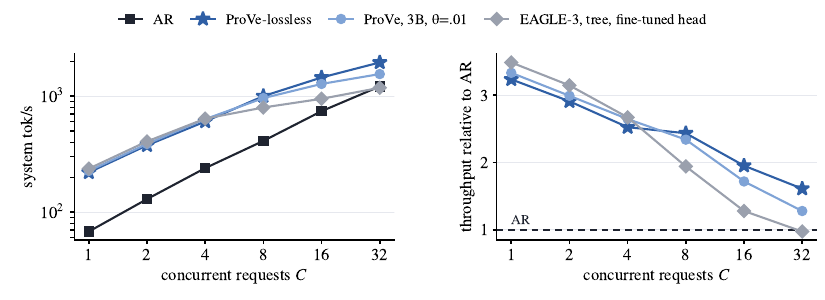}
\caption{System throughput against the number of concurrent requests $C$ on GSM8K prompts (up to 512 new tokens, about 140 generated per request), 32B instruction proposer, one two-GPU server per decoder, all four decoders on the same node. Left: generated tokens per second over all requests in flight; right: the same relative to AR at the same $C$.}
\label{fig:conc}
\end{figure}

\paragraph{Concurrency.}
Each \ours{} cycle occupies more accelerator rows per accepted token than AR decoding, free only while the device is latency bound, so the advantage narrows with the number of concurrent requests $C$ (Figure~\ref{fig:conc}).
On GSM8K prompts, throughput relative to AR at the same $C$ falls from $3.2\times$ at $C = 1$ to $2.4\times$ at $C = 8$ and $1.6\times$ at $C = 32$ for \lossless{}, and from $3.3\times$ to $2.3\times$ and $1.3\times$ for \ours{} (stock 3B verifier, $\theta = 0.01$), ahead of \lossless{} up to $C = 4$ and behind it from $C = 8$ on; the EAGLE-3 tree, at 60 rows per cycle, falls from $3.5\times$ to $1.9\times$ and reaches AR at $C = 32$.

\section{Limitations}
\label{sec:limitations}

\paragraph{Proposer training.}
The proposer requires block-specific fine-tuning and is therefore not a drop-in accelerator for an arbitrary checkpoint.
Our proposer is also substantially under-trained relative to the original \mars{} recipe because of our compute budget.
\mars{} uses five epochs of ordinary AR training followed by five epochs of block training, whereas we use only one epoch of each, or 20\% of the training in each stage.
Despite this limited training, our proposer already produces an average AR-matching prefix of 6.6 tokens on GSM8K.
More training data, longer block training, or incorporating the block-proposal objective during pretraining could substantially improve proposal quality and, consequently, the speedups of both \ours{} variants.

\paragraph{Serving implementation.}
Our implementation is research-oriented and remains less optimized than industrial inference systems.
The current serving stack incurs fixed overheads from scheduling, KV-cache management, and verifier calls that are not intrinsic to proposer-verifier decoding.
These overheads reduce the realized speedup, particularly under concurrency.
A production implementation with a more tightly integrated propose--verify loop, optimized cache management, and lower per-call overhead could translate longer committed prefixes more directly into end-to-end throughput.

\paragraph{Greedy decoding only.}
All reported experiments use greedy decoding.
\lossless{} can in principle be extended to sampling using a speculative-sampling correction that preserves the target distribution, but we have not implemented or evaluated this setting in our serving system.
The behavior and efficiency of \ours{} under sampling therefore remain to be studied.

\paragraph{\ours{} is not distribution-exact.}
Unlike \lossless{}, \ours{} with a smaller verifier uses a thresholded acceptance rule and does not preserve the proposer's greedy AR output exactly.
Its speed--accuracy tradeoff therefore depends on the verifier, threshold, and task, and we do not extrapolate the measured accuracy effects beyond the settings evaluated in this work.

\paragraph{Scope.}
Our experiments cover one model family (Qwen) and one proposer scale (32B).
The instruction proposer is evaluated on six benchmarks, while the reasoning proposer is evaluated on GSM8K and MATH-500.
Whether the same gains hold across substantially different model families and scales remains an open question.

\section{AI Use Statement}
In this work, we used generative AI tools to assist with the presentation and writing of proofs, propose or refine hypotheses, provide feedback on research methodology and experimental design, assist with method implementation, and interpret experimental results. We did not use generative AI tools to generate synthetic datasets, assist with translation, clean or reformat datasets, develop theoretical models or conceptual frameworks, formulate mathematical claims, or provide critical ingredients for proving mathematical claims. The use of generative AI for qualitative or thematic data analysis was not applicable to this work.

We also used generative AI tools to draft portions of the manuscript, improve the readability of the writing, and help identify relevant literature. All AI-assisted content was reviewed and verified by the authors. In particular, we carefully reviewed the manuscript and evaluated AI-assisted implementation using manually written unit tests. We take full responsibility for the final content of this work, including all text, claims, results, code, and other artifacts produced with the assistance of generative AI.

\end{document}